\documentclass{article} 
\usepackage{iclr2024_conference,times}

\usepackage{amsmath,amsfonts,bm}

\def\eqref#1{equation~\ref{#1}}

\def\1{\bm{1}}

\DeclareMathAlphabet{\mathsfit}{\encodingdefault}{\sfdefault}{m}{sl}
\SetMathAlphabet{\mathsfit}{bold}{\encodingdefault}{\sfdefault}{bx}{n}

\usepackage{times,graphicx,amsmath,amssymb,booktabs,tabulary,multirow,overpic,xcolor}
\usepackage{colortbl}
\usepackage{makecell}
\usepackage{subfig}
\usepackage{float}
\usepackage{url}
\usepackage{adjustbox}
\usepackage{multirow}
\usepackage{balance}
\usepackage{wrapfig, blindtext} 
\usepackage{natbib}

\definecolor{sh_gray}{rgb}{0.84,0.84,0.84}
\definecolor{sh_gray2}{rgb}{1,0.89,0.75}
\definecolor{color3}{rgb}{0.95,0.95,0.95}
\definecolor{color4}{rgb}{0.96,0.96,0.86}
\definecolor{color5}{rgb}{0.90,0.90,0.90}

\usepackage{hyperref}
\usepackage{url}

\title{Xformer: Hybrid X-Shaped Transformer for Image Denoising}

\author{
	\hspace{-3mm}Jiale Zhang$^{1}$,\enspace Yulun Zhang$^{1}$\thanks{Corresponding authors: Yulun Zhang, yulun100@gmail.com; Linghe Kong, linghe.kong@sjtu.edu.cn.}\hspace{1.5mm}%
	\thanks{The work was mainly done when Yulun Zhang was at ETH Z\"{u}rich.},\enspace Jinjin Gu$^{2,3}$,\enspace Jiahua Dong$^{4}$,\enspace Linghe Kong$^{1*}$,\enspace Xiaokang Yang$^{1}$ \\
	\hspace{-3mm}\textsuperscript{1}Shanghai Jiao Tong University,\enspace \textsuperscript{2}Shanghai AI Laboratory,\enspace \textsuperscript{3}University of Sydney,\\ \hspace{-4.5mm}\enspace \textsuperscript{4}Shenyang Institute of Automation, Chinese Academy of Sciences
}

\iclrfinalcopy 
\begin{document}
	\maketitle
	
	\vspace{-7mm}
	\begin{abstract}
		\vspace{-2mm}
		In this paper, we present a hybrid X-shaped vision Transformer, named Xformer, which performs notably on image denoising tasks. We explore strengthening the global representation of tokens from different scopes. In detail, we adopt two types of Transformer blocks. The spatial-wise Transformer block performs fine-grained local patches interactions across tokens defined by spatial dimension. The channel-wise Transformer block performs direct global context interactions across tokens defined by channel dimension. Based on the concurrent network structure, we design two branches to conduct these two interaction fashions. Within each branch, we employ an encoder-decoder architecture to capture multi-scale features. Besides, we propose the Bidirectional Connection Unit (BCU) to couple the learned representations from these two branches while providing enhanced information fusion. The joint designs make our Xformer powerful to conduct global information modeling in both spatial and channel dimensions. Extensive experiments show that Xformer, under the comparable model complexity, achieves state-of-the-art performance on the synthetic and real-world image denoising tasks. We also provide code and models at~\url{https://github.com/gladzhang/Xformer}.
		
	\end{abstract}
	
	\setlength{\abovedisplayskip}{2pt}
	\setlength{\belowdisplayskip}{2pt}

	\vspace{-6mm}
	\section{Introduction}
	\vspace{-2mm}
	\label{sec:intro}
	
	As a fundamental vision task, image denoising aims to recover the high-quality image from its noisy counterpart. It has been a very challenging problem as the denoising process is hard to distinguish the tiny textures and details from the noise. Recently, deep convolutional neural networks (CNNs) have shown great power to solve this inverse problem~\citep{zhang2017beyonddncnn,zhang2020rdnir,tian2020imageBRDNet,zhang2021plugDRUNet,hu2021pseudoP3AN}. With the help of convolution operations, deep features can be extracted to provide powerful image representations. However, the disadvantages of convolution are also obvious. Due to the poor receptive field scaling, CNN has limited ability to capture long-range dependencies among visual elements. Moreover, the convolution filters are parameter-dependent and content-independent and thus experience difficulty to show flexibility for the dynamic inputs. To address the above shortcomings, several recent works investigate the self-attention (SA) mechanism to replace the convolution and build the Transformer-based networks~\citep{swinir2021,chen2021preIPT,restormer2022,wang2022uformer,lee2022knn}.
	
	Transformer has shown state-of-the-art performance on high-level vision tasks~\citep{xu2021coat,ali2021xcit,zhang2021rest,swintransformer2021,chu2021twins,pvt2021,xie2021segformer}. The SA mechanism has great power to capture content-dependent global representations while modeling long-distance relationships. Despite of the growing computational cost, researchers are investigating the employment of Transformer in solving low-level vision problems~\citep{swinir2021,chen2021preIPT}. \citet{swinir2021} proposed SwinIR based on Swin Transformer~\citep{swintransformer2021} to utilize spatial-wise window-based SA blocks. The tokens are extracted from a square location. \citet{restormer2022} proposed Restormer to apply SA across channel dimension rather than the spatial dimension. It is demonstrated that the channel-wise SA is able to model global connectivity. For further analysis, these two types of SA mechanisms are considered to focus on different respects of global information modeling. In detail, the spatial-wise SA is good at capturing local patch-level information and modeling fine-detailed spatial features. On the other hand, the channel-wise SA is capable of capturing global channel-level information and modeling specific channel features. Especially, both types of information modeling are important for enhancing representation learning in Transformer.
	\begin{figure*}[t]
		\vspace{-4mm}
		\centering
		\begin{tabular}{c}
			\hspace{-3mm}
			\includegraphics[width=\linewidth]{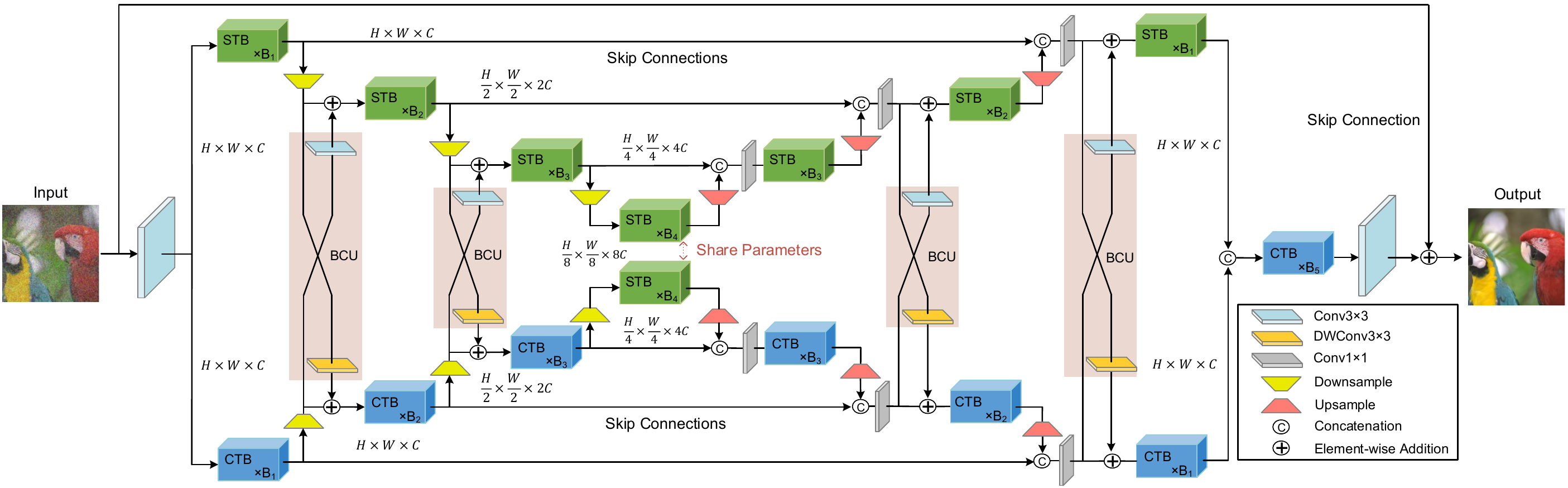} \\
		\end{tabular}
		\vspace{-3mm}
		\caption{Architecture of our proposed Xformer. The modules include spatial-wise Transformer block (STB), channel-wise Transformer block (CTB), and bidirectional connection unit (BCU).}
		\label{fig:framework}
		\vspace{-7mm}
	\end{figure*}
	\begin{wrapfigure}{r}{0.50\linewidth}
		\vspace{-2mm}
		\centering
		\resizebox{0.50\textwidth}{!}{
			\begin{tabular}{cc}
				\hspace{-0.45cm}
				\begin{adjustbox}{valign=t}
					\begin{tabular}{c}
						\includegraphics[width=0.338\textwidth]{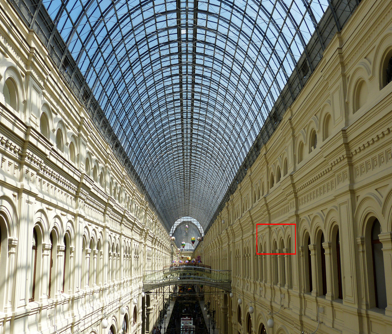}
						\\
						Urban100: img\_008
					\end{tabular}
				\end{adjustbox}
				\hspace{-0.46cm}
				\begin{adjustbox}{valign=t}
					\begin{tabular}{cccc}
						\includegraphics[width=0.249\textwidth]{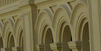} \hspace{-4mm} &
						\includegraphics[width=0.249\textwidth]{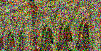} \hspace{-4mm} &
						\includegraphics[width=0.249\textwidth]{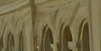} \hspace{-4mm}
						\\
						HQ  \hspace{-4mm} &
						Noisy ($\sigma$=50) \hspace{-4mm} &
						RDN \hspace{-4mm} &
						\\
						\includegraphics[width=0.249\textwidth]{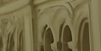} \hspace{-4mm} &
						\includegraphics[width=0.249\textwidth]{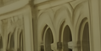} \hspace{-4mm} &
						\includegraphics[width=0.249\textwidth]{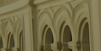} \hspace{-4mm}  
						\\ 
						SwinIR  \hspace{-4mm} &
						Restormer \hspace{-4mm} &
						\textbf{Xformer (ours)}\hspace{-4mm}
						\\
					\end{tabular}
				\end{adjustbox}
				
		\end{tabular}}
		\vspace{-2mm}
		\caption{Visual examples for Gaussian color image denoising with noise level $\sigma$=50 on Urban100.}
		\label{fig:first compare}
		\vspace{-3mm}
	\end{wrapfigure}
	\vspace{-4mm}
	
	Therefore, we explore adopting spatial-wise SA and channel-wise SA together in this paper. It remains challenging as there are gaps between these two types of SA mechanisms. We consider designing a concurrent network with dual branches. Similar parallel ideas have also been investigated for other visual tasks in recent years~\citep{chen2022mixformer,pan2022integration,peng2021conformer}. In special, the concurrent structure is beneficial for the network to build direct interactions between different branches. Furthermore, we apply spatial-wise Transformer block and channel-wise Transformer block in respective branch. Following previous works~\citep{yue2020dual,cheng2021nbnet,zamir2021multi,restormer2022}, we employ an encoder-decoder structure within each branch to obtain multi-scale features. In short, the spatial-wise branch can perform fine-grained local patches interactions across spatial-dimension tokens. The channel-wise branch can perform direct global context interactions across channel-dimension tokens.
	
	For further investigation, the concurrent network enables dual branches to model patch-level and channel-level information respectively. However, there are still some limitations. The direct concatenating operation in the end fails to effectively use these two types of features. In this way, each branch cannot capture information from different levels. Therefore, we propose the Bidirectional Connection Unit (BCU) as the bridge between two branches, which provides information fusion in an interactive manner. With BCU, the network can couple two styles of deep features. In detail, we utilize convolution layers with a $3$$\times$$3$ kernel to refine the learned deep features in corresponding branches. Then, we add the refined features to respective branches. Such a fusion operation can greatly enhance the global representation of tokens from different dimensions.
	
	Based on the designs above, we present a hybrid X-shaped Transformer for image denoising, named Xformer, as shown in Fig.~\ref{fig:framework}. We design a concurrent network with two branches. Specifically, we separately utilize spatial-wise SA blocks and channel-wise SA blocks in respective branch. Besides, we employ the proposed BCU to bridge these two branches for information fusion. The joint designs enable our network to obtain stronger global representations in Transformer. More details can be found in Sec.~\ref{sec:method}. Our Xformer can achieve superior results against recent leading image denoising methods. As shown in Fig.~\ref{fig:first compare}, our proposed method obtains visually pleasing results while others suffer from the loss of details.
	Overall, we summarize our main contributions as follows:
	\begin{itemize}
		\item We propose Xformer, an X-shaped Transformer with hybrid implementations of spatial-wise and channel-wise Transformer blocks, thereby exploiting the stronger global representation of tokens in Transformer-based neural network.
		\vspace{-1mm}
		\item We propose the Bidirectional Connection Unit (BCU) that is able to effectively couple the learned representations from two branches of Xformer. This simple design significantly enhances the global information modeling of our method.
		\vspace{-1mm}
		\item We employ Xformer to train an efficient and effective Transformer-based network for image denoising. We conduct extensive experiments on the synthetic and real-world noise removal tasks. Our method can achieve state-of-the-art performance.
	\end{itemize}
	
	\section{Related Work}
	\label{sec:related_work}
	\noindent \textbf{Image Denoising.} Due to the powerful generalizing ability from large-scale data, CNN-based methods~\citep{zhang2020rdnir,tian2020imageBRDNet,zhang2021plugDRUNet,hu2021pseudoP3AN} have achieved superior performance over the traditional denoising algorithms~\citep{perona1990scale,mairal2009non,elad2006image}. \citet{zhang2017beyonddncnn} proposed DnCNN as a representative image denoising method, which trained mappings from noisy images to noises. For further improvements, many subsequent works used more elaborate network architecture designs, including encoder-decoder structure~\citep{yue2020dual,cheng2021nbnet,zamir2021multi}, non-local modules~\citep{liu2018nonNLRN,zhang2019rnan}, attention mechanism~\citep{zhang2021accurate}, and dynamic convolution~\citep{jiang2022fast}. Unfortunately, most CNNs suffer from the limited ability to model long-range dependencies, which is crucial for recovering clean images. Very recently,  researchers have started to utilize self-attention strategy to replace the single convolution operation.~\citep{swinir2021,chen2021preIPT,restormer2022,wang2022uformer,lee2022knn}.
	
	\vspace{0.4em}
	\noindent \textbf{Vision Transformer.} In recent years, Transformer has achieved impressive success in machine translation tasks~\citep{vaswani2017attention}. It also performs outstandingly to solve numerous high-level vision problems~\citep{zhang2021rest,chu2021twins,pvt2021,xie2021segformer} due to the content-dependent global receptive field of the network. \citet{dosovitskiy2020image} firstly proposed ViT to introduce Transformer into image recognition. Inspired by these, more and more works started to apply Transformer to solve low-level vision tasks~\citep{swinir2021,chen2021preIPT,wang2022uformer,restormer2022}. \citet{wang2022uformer} proposed a general U-shaped Transformer named Uformer based on U-Net~\citep{ronneberger2015u} for image restoration. \citet{restormer2022} proposed a strong baseline model named Restormer and achieved state-of-the-art performance in several image restoration tasks. \citet{chen2021preIPT} proposed IPT to apply standard Transformer blocks while using pre-training on additional datasets. To sum up, these works have not explored enhancing the global representation of tokens from different dimensions. In contrast, we design a general X-shaped Transformer to bridge this gap.
	
	\vspace{0.4em}
	\noindent \textbf{Concurrent Network.} Compared to the commonly-used serial network, the concurrent network has parallel branches in the whole network architecture. Therefore, it has the natural advantage of simultaneously conducting different types of representation learning and building direct interactions between dual branches. Recently, few efforts have been made to explore this field. Some works only focused on parallel blocks such as \citet{chen2022mixformer} and \citet{pan2022integration}. \citet{peng2021conformer} defined a representative concurrent network named Conformer to solve some high-level vision problems. They designed two branches to respectively leverage convolution operators and self-attention mechanisms. Specifically, the CNN branch preserves fine-detailed local features. The Transformer branch captures long-range dependencies. In this paper, we also design a concurrent network. Different from it, our proposed Xformer applies different Transformer blocks in two branches and we focus on low-level vision tasks. Besides, we propose the Bidirectional Connection Unit (BCU) to greatly enhance the information fusion within two branches.
	
	\vspace{-3mm}
	\section{Method}
	\label{sec:method}
	\vspace{-3mm}
	\subsection{Overall Pipeline}
	\vspace{-1mm}
	As shown in Fig.~\ref{fig:framework}, our Xformer is a hybrid X-shaped Transformer-based network with two branches. Following the design of U-Net structures~\citep{cheng2021nbnet,restormer2022,wang2022uformer}, each branch is treated as a separate U-shaped network with skip connections between encoders and decoders. We utilize the spatial-wise window-based Transformer blocks (STBs) to construct the spatial-wise branch. The channel-wise cross-covariance Transformer blocks (CTBs) are used to construct the channel-wise branch. Then, we design the Bidirectional Connection Unit (BCU) to bridge the dual branches for feature complementarity. It can bring information fusion for different branches. Besides, we provide two additional designs. Firstly, we make the last encoder involving STBs of two branches share parameters for the purpose of computational efficiency. Secondly, we concatenate the output features from two branches and send them to a new refinement module involving CTBs. The overall pipeline is as follows.
	
	Given a degraded image $\textrm{$\textbf{I}$}$ $\in$ $\mathbb{R}^{H\times W \times C_{in}}$ where $H$, $W$, and $C_{in}$ are the height, width, and input channels, our proposed Xformer first uses a 3$\times$3 convolutional layer (Conv) to obtain the shallow feature $\textrm{$\textbf{F}_0$}$ $\in$ $\mathbb{R}^{H\times W \times C}$, where $C$ is the size of new feature dimension. Next, the feature $\textrm{$\textbf{F}_0$}$ is concurrently sent to two 4-level symmetric encoder-decoder branches. Through these two branches, it is transformed into two new deep features $\textrm{$\textbf{F}_s$}$, $\textrm{$\textbf{F}_c$}$ $\in$ $\mathbb{R}^{H\times W \times C}$. In detail, each encoder or decoder contains cascaded Transformer blocks. Encoders take $\textrm{$\textbf{F}_0$}$ as input and reduce half of the spatial resolution while doubling the number of feature channels as the stage grows. Decoders take the low-resolution features as input and reduce half of the feature channels while doubling the size of feature maps. By counting, input features experience three times of downsampling and upsampling. Besides, the up-sampled features are concatenated with the corresponding features from encoders via skip connections for the recovery improvement. The concatenated features further pass through a 1$\times$1 Conv to reduce channels by half. For information fusion, the down-sampled features after the first and second encoders pass through the BCU and then integrate to the opposite branch through element-wise add. Similarly, the reduced features after the first and second decoders pass through the BCU and are added to the opposite branch. In the end, the new features $\textrm{$\textbf{F}_s$}$ and $\textrm{$\textbf{F}_c$}$ are concatenated and then flow to the refinement module. The output is further transmitted into a 3$\times$3 Conv to obtain a residual image $\textrm{$\textbf{I}_r$}$ $\in$ $\mathbb{R}^{H\times W \times C_{in}}$. Finally, the restored image is generated by $\textrm{$\hat{\textbf{I}}$} = \textrm{$\textbf{I}$} + \textrm{$\textbf{I}_r$}$.
	
	\begin{figure*}[t]
		\centering
		\vspace{-3mm}
		\begin{tabular}{c}
			\hspace{-3mm}
			\includegraphics[width=\linewidth]{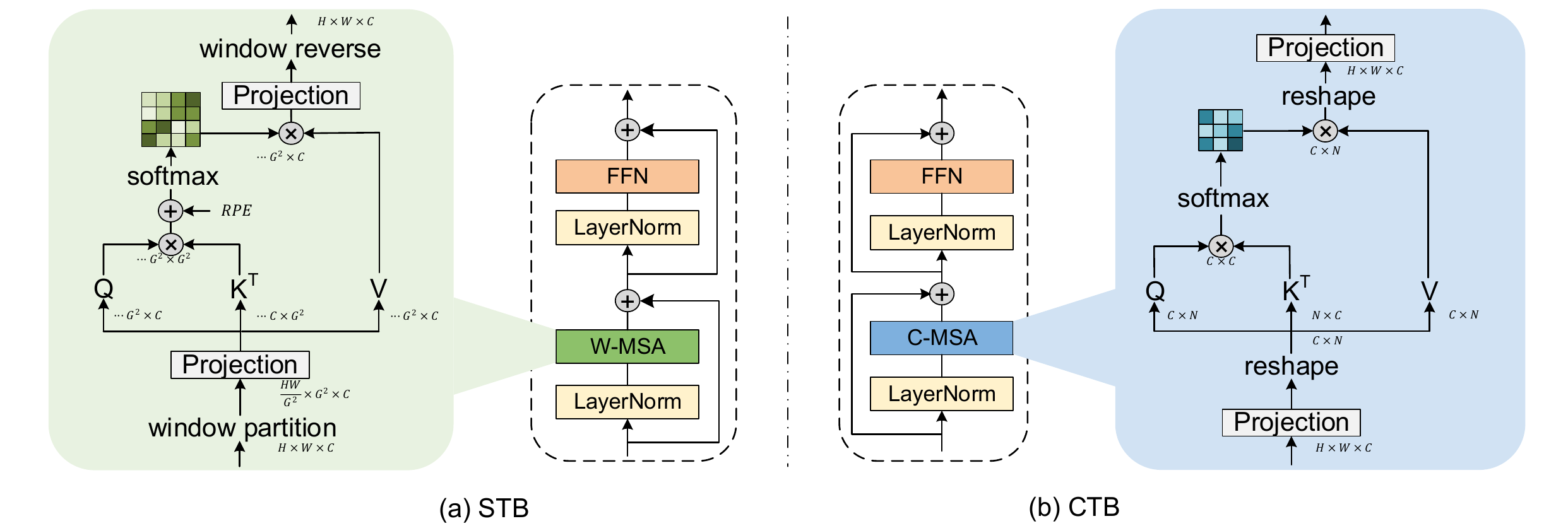} \\
		\end{tabular}
		\vspace{-4mm}
		\caption{Module architectures of spatial-wise and channel-wise Transformer blocks (STB\texttt{|}CTB).}
		\label{fig:transformer}
		\vspace{-8mm}
	\end{figure*}
	
	\vspace{-3mm}
	\subsection{Dual Branches}
	\vspace{-3mm}
	\noindent \textbf{Spatial-wise Branch.} As shown in the upper part of Fig.~\ref{fig:framework}, the spatial-wise branch adopts the encoder-decoder structure with skip connections. The components of encoders and decoders are cascaded spatial-wise window-based Transformer blocks (STBs). We provide the details of STB in Fig.~\ref{fig:transformer}\textcolor{red}{(a)}. We make $\textrm{$\textbf{x}_l$}$ as the output at the $l_{th}$ block. We formulate the calculation process of STB as
	\begin{equation}
		\label{equ:attention spatial-1}
		\begin{aligned}
			\textrm{$\textbf{x}{'}_l$} &= \operatorname{W-MSA}(\operatorname{LN}(\textrm{$\textbf{x}_{l-1}$}))+\textrm{$\textbf{x}_{l-1}$}, \\
			\textrm{$\textbf{x}_l$} &= \operatorname{FFN}(\operatorname{LN}(\textrm{$\textbf{x}{'}_l$}))+\textrm{$\textbf{x}{'}_l$},
		\end{aligned}
	\end{equation}
	where W-MSA means the window-based multi-head self-attention. Here we assume that the number of heads is $1$ to transfer MSA to singe-head mode. Given the feature $\textrm{$\textbf{X}$}$ $\in$ $\mathbb{R}^{H\times W \times C}$ generated by layer normalization (LN)~\citep{ba2016layer}, W-MSA first split it into non-overlapping $G$$\times$$G$ windows to get features $\textrm{$\textbf{X}^i$}$ $\in$ $\mathbb{R}^{G^2 \times C}$ for $i_{th}$ window. Next, it performs linear projecting to generate query ($\textrm{$\textbf{Q}^i$}$), key ($\textrm{$\textbf{K}^i$}$), and value ($\textrm{$\textbf{V}^i$}$), yielding $\mathbf{Q}^i$=$X^iW^Q$, $\mathbf{K}^i$=$X^iW^K$, and $\mathbf{V}^i$=$X^iW^V$, where $W^Q$,$W^K$,$W^V$ $\in$ $\mathbb{R}^{C \times C}$ are learnable parameters. We formulate the calculation in $i_{th}$ window as
	\begin{equation}
		\label{equ:attention spatial-2}
		\begin{split}
			\textrm{$\hat{\textbf{X}}^i$} &= \operatorname{Softmax}(\frac{\textrm{$\textbf{Q}^i$} \textrm{$\textbf{K}^i$}^T}{\sqrt{\textrm{$\textbf{C}$}}}+\textrm{$\textbf{B}$})\textrm{$\textbf{V}^i$},
		\end{split}
		\vspace{-0.9mm}
	\end{equation}
	where $\textrm{$\hat{\textbf{X}}^i$}$ is the output feature map in $i_{th}$ window and $\textrm{$\textbf{B}$}$ is the relative position encoding (RPE)~\citep{shaw2018self,raffel2020exploring}. Furthermore, the features from all windows are projected together and reshaped to the new feature map of size $\mathbb{R}^{H\times W \times C}$, as the last output of W-SMA. For the Feed-forward network (FFN), we use the basic multi-layer perception (MLP) used in recent works~\citep{swintransformer2021,swinir2021} to deal with the input features. In short, the STB utilizes non-overlapping windows to generate shorter token sequences for the self-attention computation, which can enable the network to obtain fine-grained local patches interactions. 
	
	\vspace{.2mm}
	\noindent \textbf{Channel-wise Branch.} Similarly, this channel-wise branch contains a 4-level encoder-decoder structure. In special, the encoders and decoders are constructed by cascaded channel-wise Transformer blocks (CTBs). The details of CTB are shown in Fig.~\ref{fig:transformer}\textcolor{red}{(b)}. Assuming that $\textrm{$\textbf{x}_k$}$ is the output at the $k_{th}$ block, the calculation process can be formulated as
	\begin{equation}
		\label{equ:attention channel-1}
		\begin{aligned}
			\textrm{$\textbf{x}{'}_k$} &= \operatorname{C-MSA}(\operatorname{LN}(\textrm{$\textbf{x}_{k-1}$}))+\textrm{$\textbf{x}_{k-1}$}, \\
			\textrm{$\textbf{x}_k$} &= \operatorname{FFN}(\operatorname{LN}(\textrm{$\textbf{x}{'}_k$}))+\textrm{$\textbf{x}{'}_k$},
		\end{aligned}
	\end{equation}
	where C-MSA means the channel-wise multi-head self-attention. We also assume that the number of heads is $1$ and transfer MSA to a singe-head fashion. Given the normalized feature $\textrm{$\textbf{X}$}$ $\in$ $\mathbb{R}^{H\times W \times C}$, C-MSA first utilizes the projecting module to get prepared query, key and value. In order to introduce contextualized information into self-attention computation, we choose to use 3$\times$3 depth-wise convolution (Conv) following 1$\times$1 Conv to generate query ($\textrm{$\textbf{Q}$}$), key ($\textrm{$\textbf{K}$}$), and value ($\textrm{$\textbf{V}$}$). We make $\mathbf{Q}$=$W^Q_{d}W^Q_{p}X$, $\mathbf{K}$=$W^K_{d}W^K_{p}X$, and $\mathbf{V}$=$W^V_{d}W^V_{p}X$, where $W^{(\cdot)}_{p}$ means parameters of 1$\times$1 point-wise Conv and $W^{(\cdot)}_{d}$ means parameters of 3$\times$3 depth-wise Conv. Then, the obtained $\textrm{$\textbf{Q, K, V}$}$ are reshaped into new feature maps of size $\mathbb{R}^{C \times N}$, where $\mathbf{N}$=$H$$\times$$W$. The query and key are further normalized to prepare for cross-covariance attention. The new transposed attention map is calculated by $\textrm{$\textbf{Q}$}$ and $\textrm{$\textbf{K}$}^T$ with size of $\mathbb{R}^{C \times C}$. The calculation process of the C-MSA is formulated as
	\begin{equation}
		\label{equ:attention channel-2}
		\begin{split}
			\textrm{$\hat{\textbf{X}}$} &= \operatorname{Softmax}(\textrm{$\textbf{Q}$} \textrm{$\textbf{K}$}^T / \tau)\textrm{$\textbf{V}$},
		\end{split}
	\end{equation}
	where $\tau$ is a learnable temperature parameter and $\textrm{$\hat{\textbf{X}}$}$ $\in$ $\mathbb{R}^{C \times N}$ is the output. Then $\textrm{$\hat{\textbf{X}}$}$ is reshaped to the original feature size of $\mathbb{R}^{H \times W \times C}$. The output further passes through a linear projecting layer. Added by the shortcut $\textrm{$\textbf{x}_{k-1}$}$, new features $\textrm{$\textbf{x}{'}_k$}$ is transmitted to the following part. For the FFN, we introduce the gating mechanism and depth-wise convolutions proposed in the recent work~\citep{restormer2022} to enrich the feature transferring, which is validated to be effective. Equipped with the used C-MSA and FFN, the CTB enjoys strong ability to capture direct global context interactions.
	
	\vspace{-3mm}
	\subsection{Bidirectional Connection Unit} \label{subsec: bcu}
	\noindent \textbf{Motivation.} As the dual branches enable the network to capture both patch-level and channel-level information, the information fusion is treated as an important step to enhance global information modeling. Simple concatenating operation is not able to effectively utilize information from different branches. The direct connection of dual branches is not the best choice. Therefore, we propose the Bidirectional Connection Unit (BCU) to couple the deep features from their respective feed-forward processes for feature complementarity. We demonstrate that the proposed BCU plays an important role to provide enhanced information fusion.
	
	\vspace{0.4em}
	\noindent \textbf{Specific Design.} We design the BCU to bridge the two branches in an interactive manner. We carry out the specific feature complementarity like the form of absolute position encoding~\citep{vaswani2017attention}. On one hand, we add the global context information brought by channel-wise self-attention to the feature maps of the spatial-wise branch. On the other hand, we add the fine-grained patch-level information brought by local patches interactions to the feature maps of the channel-wise branch. In detail, the BCU contains two simple convolution layers. Specifically, we use a 3$\times$3 depth-wise convolution layer to refine the deep features from the spatial-wise branch for the purpose of saving computational consumption. We use a common 3$\times$3 convolution layer to refine features from the channel-wise branch to provide more channel-dimension interactions. With the 3$\times$3 kernel size, the feature refinement process can provide extra contextualized information. The specific implementation is shown in Fig.~\ref{fig:framework}.
	
	\vspace{-4mm}
	\subsection{Implementation Details}
	\label{sec:Implementation}
	
	\noindent \textbf{Specific Settings.} Firstly, we set the layer numbers of both branches the same, which are [2, 4, 4, 6, 4, 4, 2]. The number of CTBs in the refinement stage is set to 4. Secondly, we set the number of heads in corresponding layers to [1, 2, 4, 8, 4, 2, 1]. The head number of CTBs in the refinement stage is set to 1. Meanwhile, the channels number of shallow features generated by the first convolution layer is set to 48. The expansion size of hidden layers in FFN is set to 2.66. Thirdly, the window size in spatial-wise Transformer blocks is set to 16. Note that we also utilize the shifted-window strategy~\citep{swintransformer2021}. Besides, we use pixel-unshuffle and pixel-shuffle operations~\citep{shi2016real} for downsampling and upsampling. Lastly, following the recent work~\citep{restormer2022}, we use the progressive training strategy for fair comparisons.
	
	\vspace{0.4em}
	\noindent \textbf{Loss Function.} Following most recent works~\citep{lai2017deep,zhang2020rdnir,restormer2022}, we use $L_1$ loss function to optimize our proposed Xformer. For image denoising, the goal of training Xformer is to minimize the $L_1$ loss, which is formulated as
	\begin{equation}
		\label{equ:l1loss}
		\mathcal{L} = \lVert\hat{I}_{HQ} - I_{HQ}\rVert_1,
	\end{equation}
	where $\hat{I}_{HQ}$ is the output of our Xformer and $I_{HQ}$ is the corresponding ground-truth image.
	
	\begin{table*}[t]\vspace{-6mm}
		\centering
		\subfloat[\small  Ablation study of block setting. \label{tab:STBCTB}]{ 
			\scalebox{0.61}{
				\begin{tabular}{ c c c c}
					\toprule
					\rowcolor{color3} Method & All STB & All CTB & STB+CTB \\
					\midrule
					Params (M)  & 26.03 & 28.81 & 25.23  \\
					FLOPs (G)  & 38.1 & 42.3 & 42.2   \\
					PSNR (dB) & 29.87 & 29.67 & 29.94  \\
					SSIM    & 0.8851  & 0.8830 & 0.8865 \\
					
					\bottomrule
		\end{tabular}}}\hspace{-1mm}\vspace{0.5mm}
		\subfloat[\small Ablation study of BCU settings.\label{tab:bcu}]{
			\scalebox{0.61}{
				\begin{tabular}{c c c c c }
					\toprule
					\rowcolor{color3} Method & w/o BCU & BCU-1 & BCU-2 & Complete BCU\\
					\midrule
					Params (M)  & 24.70 & 24.71 & 25.22 & 25.23  \\
					FLOPs (G)  & 40.9 & 40.9 & 42.2  & 42.2 \\
					PSNR (dB) & 29.82 & 29.84 & 29.92 & 29.94  \\
					SSIM    & 0.8842  & 0.8848 & 0.8859 & 0.8865 \\
					
					\bottomrule
		\end{tabular}}}\hspace{-1mm}\vspace{0.5mm}
		\subfloat[\small Whether to use shift.\label{tab:shift}]{
			\scalebox{0.61}{
				\begin{tabular}{c c c c c}
					\toprule
					\rowcolor{color3} Method & w/o Shift & w/ Shift   \\
					\midrule
					Params (M) & 25.23 & 25.23 \\
					FLOPs (G) & 42.2 & 42.2 \\
					PSNR (dB) & 29.88 & 29.94 \\
					SSIM  & 0.8852  & 0.8865 \\
					
					\bottomrule
		\end{tabular}}} \vspace{-4mm}\\
		\subfloat[\small Ablation study of designed models with different branches.\label{table:branch}]{
			\scalebox{0.61}{
				\begin{tabular}{c| c c c | c | c c c c}
					\toprule
					\rowcolor{color3} ID & STB & CTB & BCU & Structure & Params (M) & FLOPs (G) &   PSNR (dB) & SSIM\\
					\midrule
					1 &\checkmark & & & single-branch & 26.48 & 40.6 & 29.84 & 0.8853\\
					2 & & \checkmark & & single-branch & 26.11 &  38.7 & 29.68 & 0.8829\\
					3 &\checkmark & \checkmark& & two-branches & 24.70 & 40.9 & 29.82 & 0.8842\\
					4 &\checkmark & \checkmark &\checkmark& two-branches & 25.23 & 42.2 & 29.94 & 0.8865\\
					\bottomrule
		\end{tabular}}}\hspace{1mm}
		\subfloat[\small Params-FLOPs-PSNR comparisons. \label{table:complexity}]{
			\scalebox{0.61}{
				\begin{tabular}{c c c c}
					\toprule
					\rowcolor{color3} Method & SwinIR & Restormer & Xformer (ours)\\
					\midrule
					Params (M) & 11.50 & 26.11 & 25.23\\
					FLOPs (G) & 201.2 & 38.7 & 42.2 \\
					McMaster & 30.22 & 30.30 & 30.38 \\
					Urban100 & 29.82 & 30.02 & 30.36 \\
					\bottomrule
					
		\end{tabular}}}
		\vspace{-5mm}
		\caption{\small Ablation experiments (a-d) and model complexity comparisons (e). For ablation, we train models on Gaussian color image denoising task with $\sigma$=50 for 100k iterations and test on Urban100.}
		\label{tab:ablations}\vspace{-2mm}
	\end{table*}
	
	\begin{figure}[t]
		\scriptsize
			\centering
			\begin{tabular}{cc}
				\hspace{-0.6cm}
				\begin{adjustbox}{valign=t}
					\begin{tabular}{cccccc}
						&
						\multicolumn{2}{c}{Features from the last CTB} \hspace{-2mm}&
						\multicolumn{2}{c}{Features from the last STB}
						\\
						\includegraphics[width=0.090\textwidth]{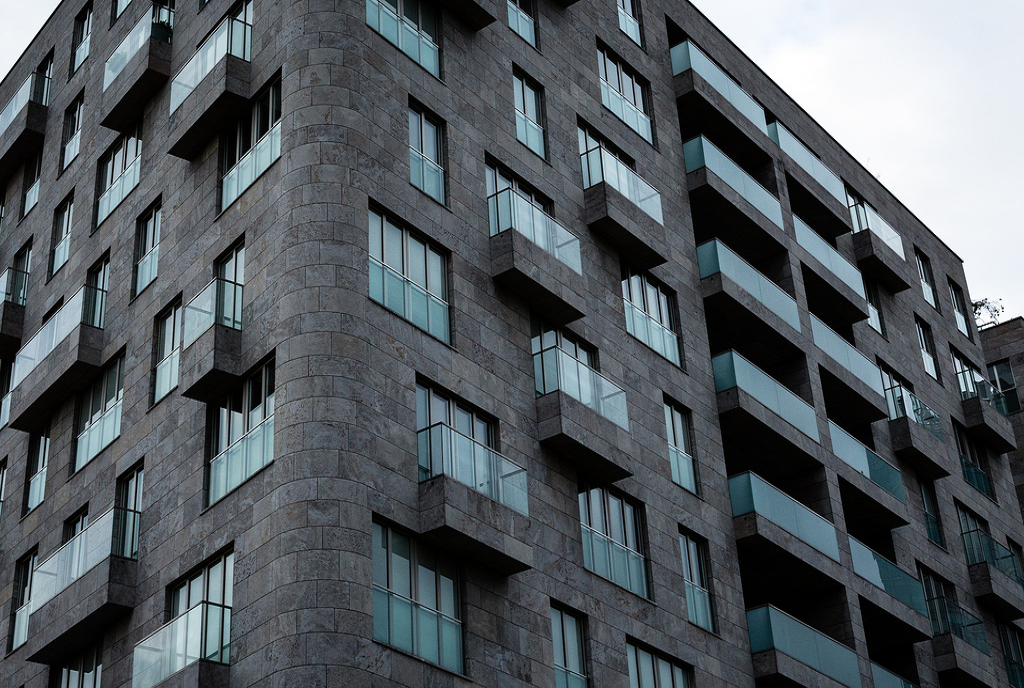} \hspace{-4mm} &
						\includegraphics[width=0.090\textwidth]{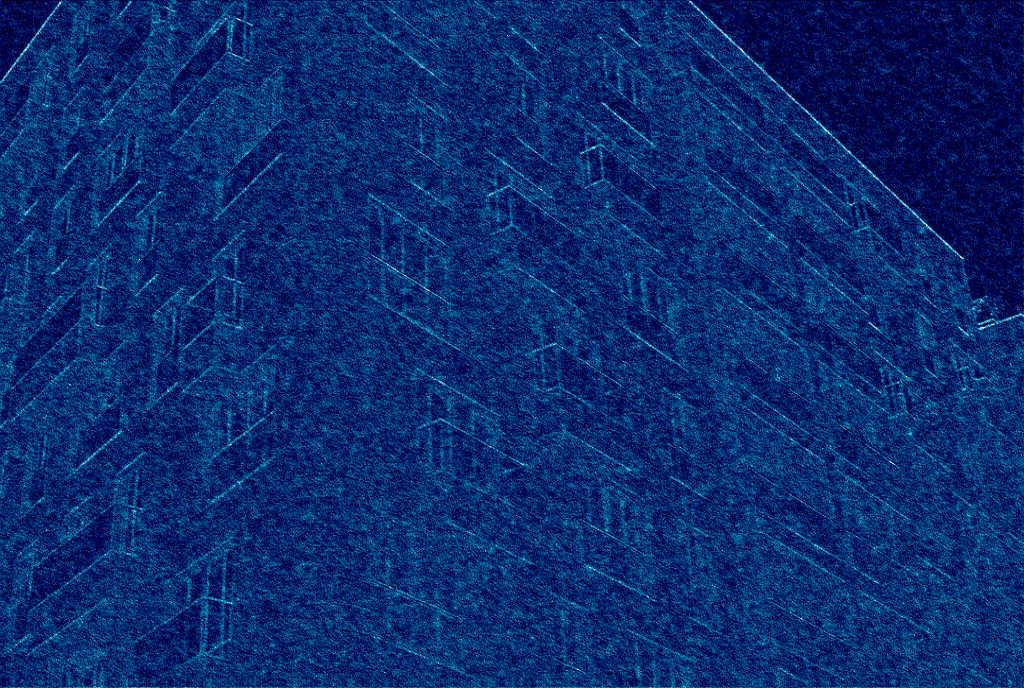} \hspace{-4mm} &
						\includegraphics[width=0.090\textwidth]{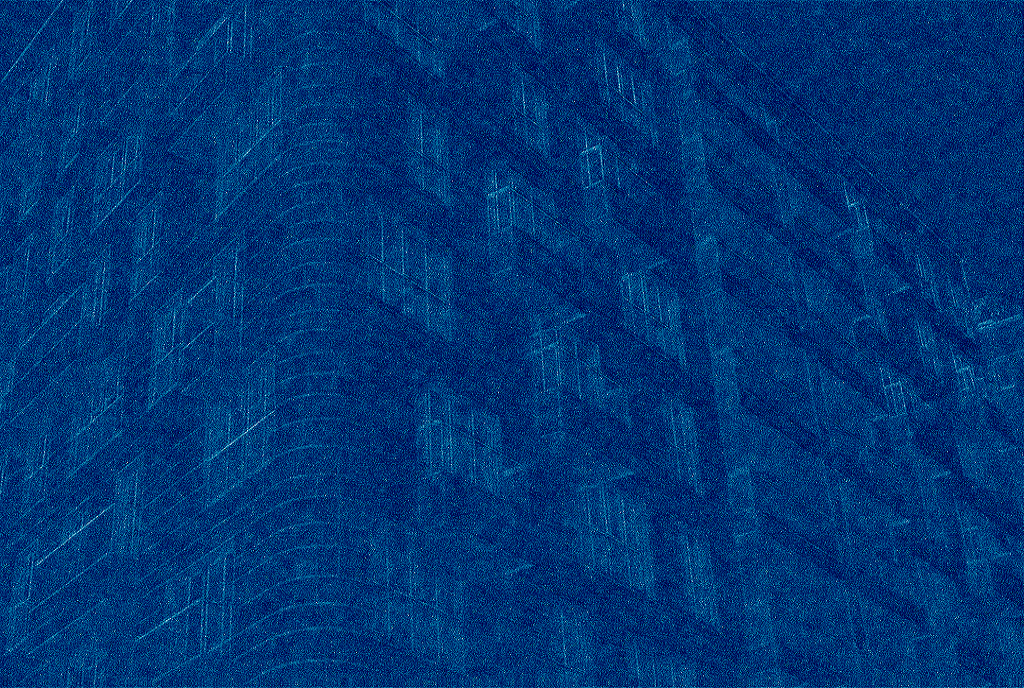} \hspace{-4mm} &
						\includegraphics[width=0.090\textwidth]{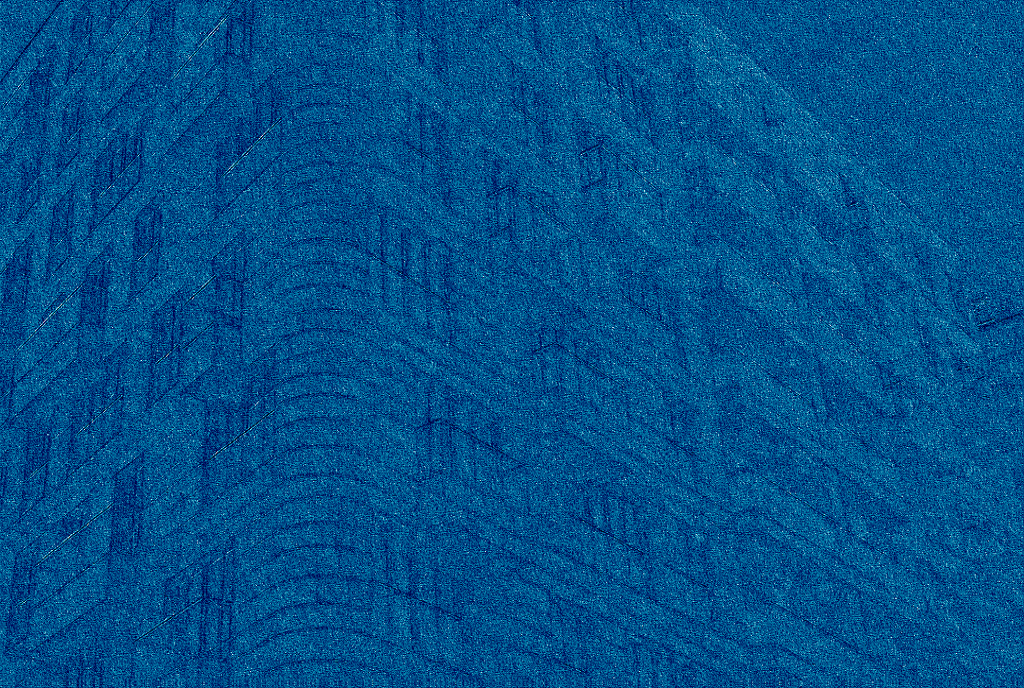} \hspace{-6mm} &
						\includegraphics[width=0.090\textwidth]{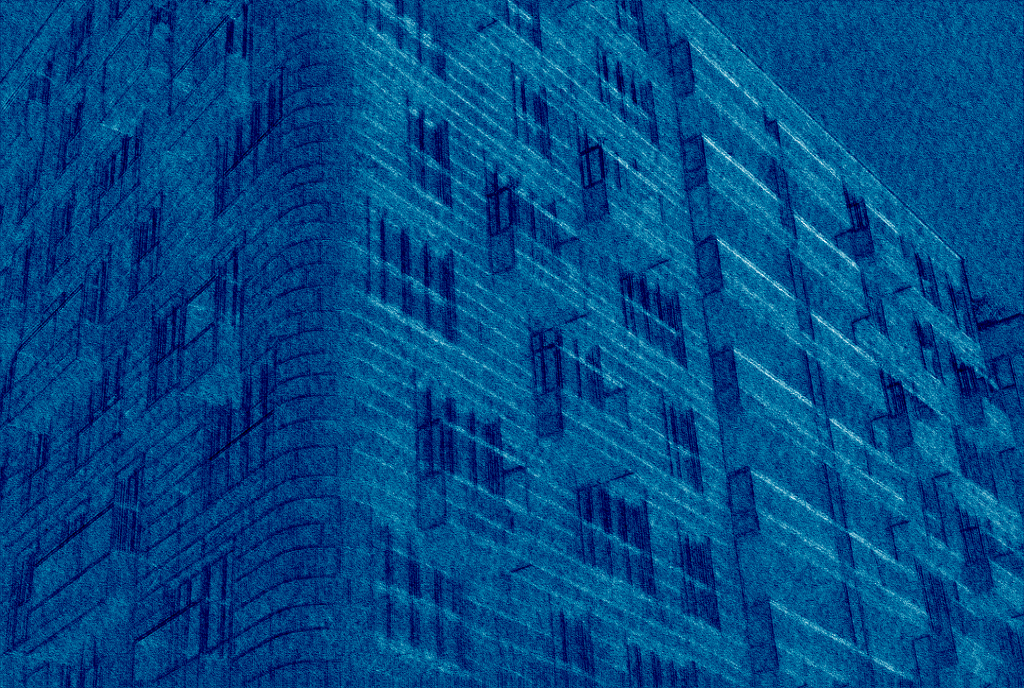} \hspace{-4mm} &
						\\
						Urban100\_HR \hspace{-4mm} &
						w/o BCU\hspace{-4mm} &
						w/ BCU  \hspace{-4mm} &
						w/o BCU \hspace{-4mm} &
						w/ BCU \hspace{-4mm} &
						\\
					\end{tabular}
				\end{adjustbox}
				\hspace{-0.8cm}
				\begin{adjustbox}{valign=t}
					\begin{tabular}{cccccc}
						&
						\multicolumn{2}{c}{Features from the last CTB} \hspace{-2mm}&
						\multicolumn{2}{c}{Features from the last STB}
						\\
						\includegraphics[width=0.090\textwidth]{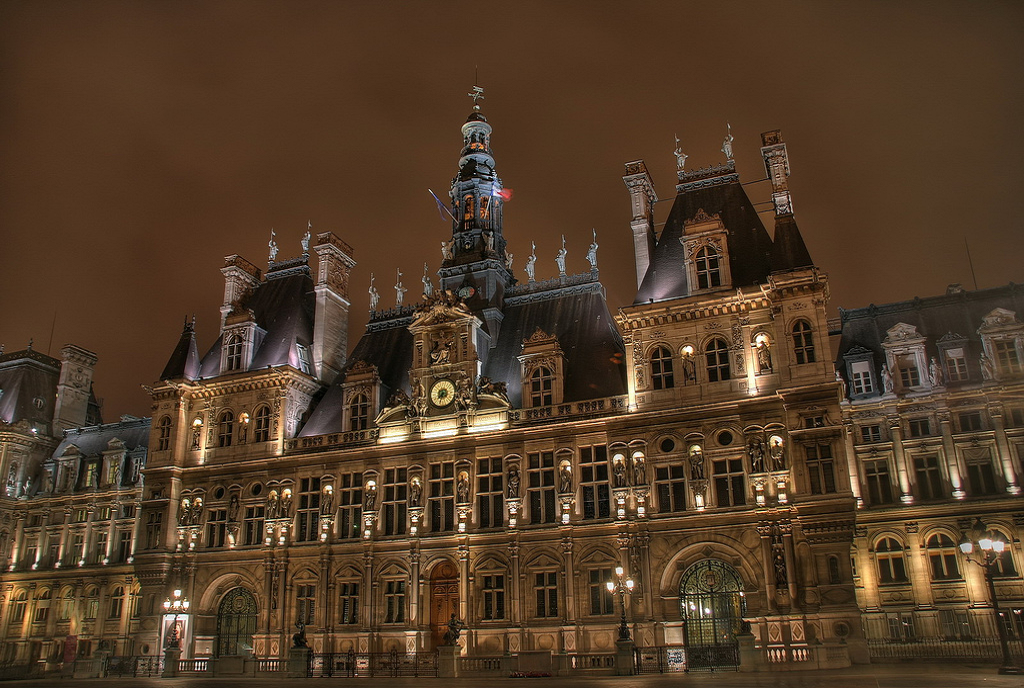} \hspace{-4mm} &
						\includegraphics[width=0.090\textwidth]{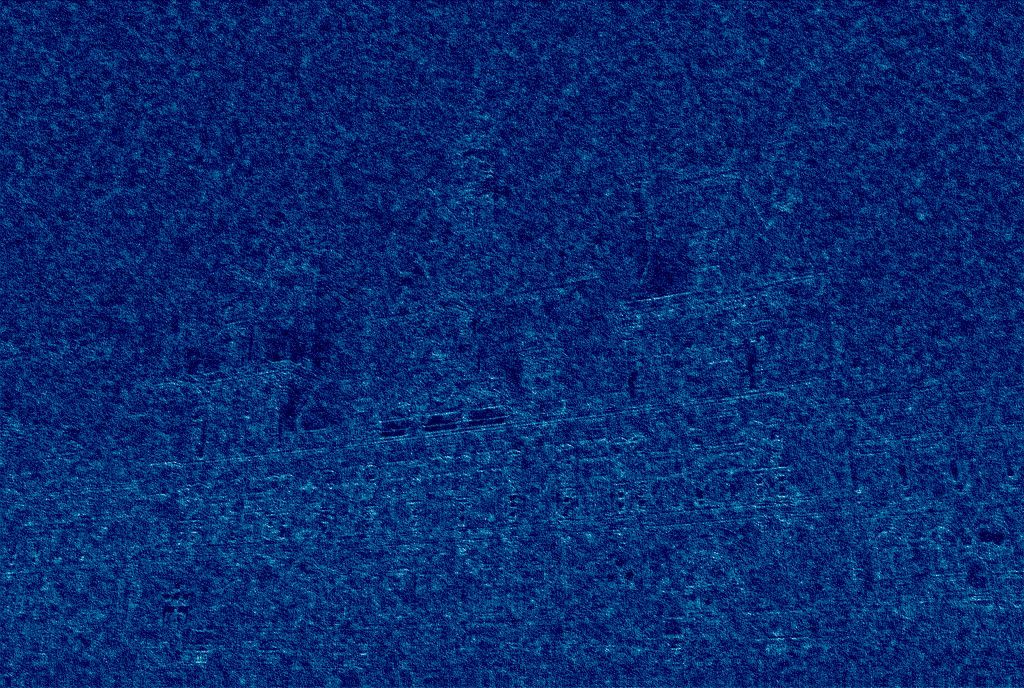} \hspace{-4mm} &
						\includegraphics[width=0.090\textwidth]{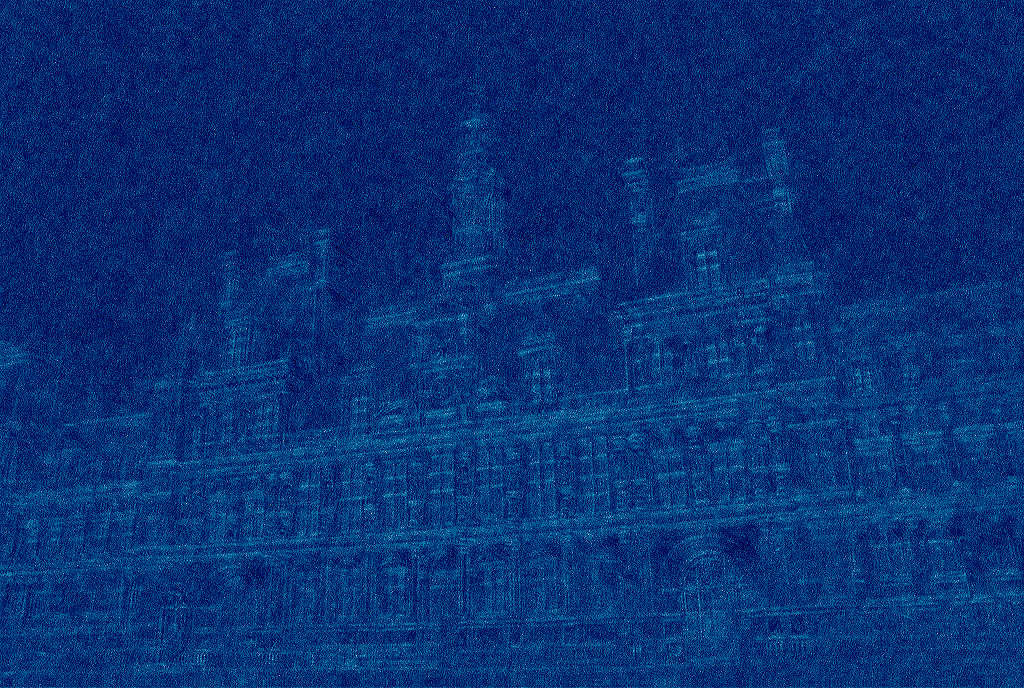} \hspace{-4mm} &
						\includegraphics[width=0.090\textwidth]{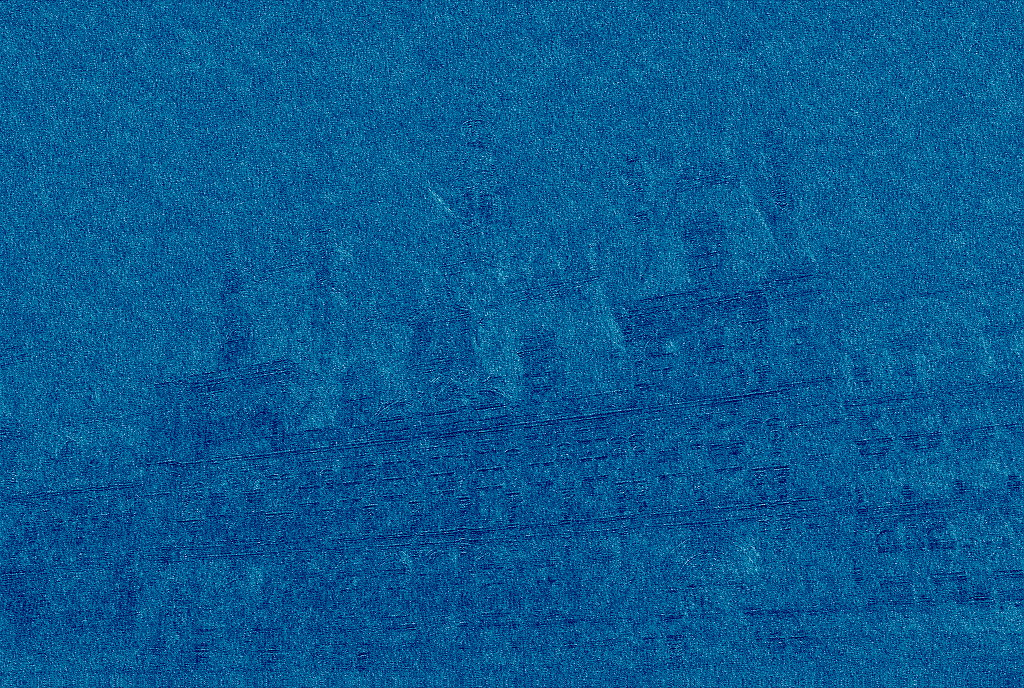} \hspace{-6mm} &
						\includegraphics[width=0.090\textwidth]{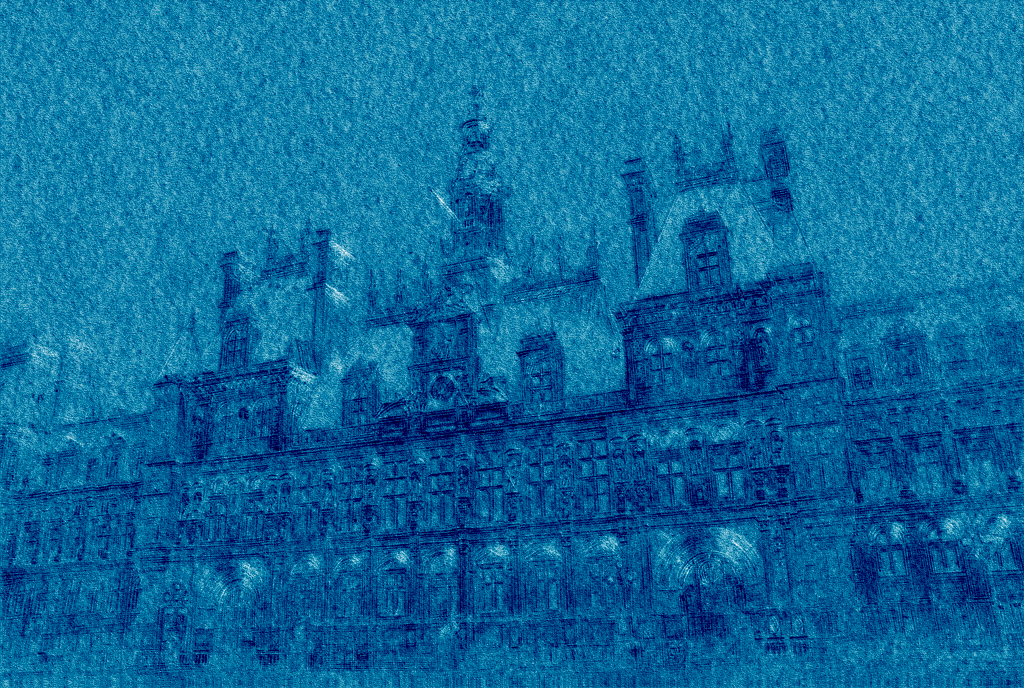} \hspace{-4mm} &
						\\
						Urban100\_HR \hspace{-4mm} &
						w/o BCU\hspace{-4mm} &
						w/ BCU  \hspace{-4mm} &
						w/o BCU \hspace{-4mm} &
						w/ BCU \hspace{-4mm} &
						\\
					\end{tabular}
				\end{adjustbox}
				\\
			\end{tabular}
		\vspace{-2mm}
		\caption{Visualization of feature maps from the last STB and CTB in the encoder-decoder module of dual branches. We compare different situations about whether to use BCU.}
		\vspace{-4mm}
		\label{fig:feature visual}
	\end{figure}
	
	\begin{figure*}[t]
		\scriptsize
		\centering
		\vspace{-5mm}
		\resizebox{1\textwidth}{!}{
			\begin{tabular}{cc}
				\hspace{-0.45cm}
				\begin{adjustbox}{valign=t}
					\begin{tabular}{c}
						\includegraphics[width=0.213\textwidth]{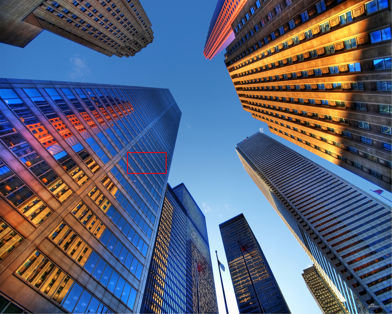}
						\\
						Urban100: img\_012
					\end{tabular}
				\end{adjustbox}
				\hspace{-0.46cm}
				\begin{adjustbox}{valign=t}
					\begin{tabular}{cccccc}
						\includegraphics[width=0.149\textwidth]{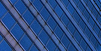} \hspace{-4mm} &
						\includegraphics[width=0.149\textwidth]{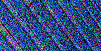} \hspace{-4mm} &
						\includegraphics[width=0.149\textwidth]{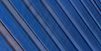} \hspace{-4mm} &
						\includegraphics[width=0.149\textwidth]{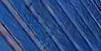} \hspace{-4mm} &
						\includegraphics[width=0.149\textwidth]{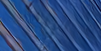}
						\hspace{-4mm} 
						\\
						HQ  \hspace{-4mm} &
						Noisy ($\sigma$=50) \hspace{-4mm} &
						BM3D \hspace{-4mm} &
						IRCNN \hspace{-4mm} &
						DnCNN \hspace{-4mm}
						\\
						\includegraphics[width=0.149\textwidth]{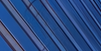} \hspace{-4mm} &
						\includegraphics[width=0.149\textwidth]{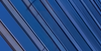} \hspace{-4mm} &
						\includegraphics[width=0.149\textwidth]{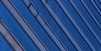} \hspace{-4mm} &
						\includegraphics[width=0.149\textwidth]{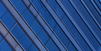} \hspace{-4mm} &
						\includegraphics[width=0.149\textwidth]{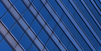} \hspace{-4mm}  
						\\ 
						RNAN \hspace{-4mm} &
						RDN \hspace{-4mm} &
						SwinIR \hspace{-4mm} &
						Restormer \hspace{-4mm} &
						\textbf{Xformer (ours)}\hspace{-4mm}
						\\
					\end{tabular}
				\end{adjustbox}
				\\
				\hspace{-0.42cm}
				\begin{adjustbox}{valign=t}
					\begin{tabular}{c}
						\includegraphics[width=0.216\textwidth]{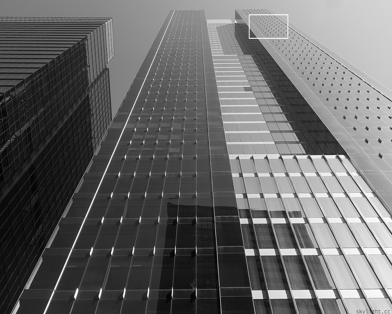}
						\\
						Urban100: img\_033
					\end{tabular}
				\end{adjustbox}
				\hspace{-0.46cm}
				\begin{adjustbox}{valign=t}
					\begin{tabular}{cccccc}
						\includegraphics[width=0.149\textwidth]{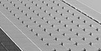} \hspace{-4mm} &
						\includegraphics[width=0.149\textwidth]{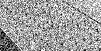} \hspace{-4mm} &
						\includegraphics[width=0.149\textwidth]{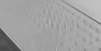} \hspace{-4mm} &
						\includegraphics[width=0.149\textwidth]{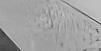} \hspace{-4mm} &
						\includegraphics[width=0.149\textwidth]{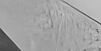}
						\hspace{-4mm} 
						\\
						HQ\hspace{-4mm} &
						Noisy ($\sigma$=50) \hspace{-4mm} &
						BM3D \hspace{-4mm} &
						IRCNN \hspace{-4mm} &
						DnCNN \hspace{-4mm}
						\\
						\includegraphics[width=0.149\textwidth]{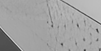} \hspace{-4mm} &
						\includegraphics[width=0.149\textwidth]{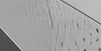} \hspace{-4mm} &
						\includegraphics[width=0.149\textwidth]{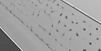} \hspace{-4mm} &
						\includegraphics[width=0.149\textwidth]{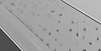} \hspace{-4mm} &
						\includegraphics[width=0.149\textwidth]{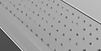} \hspace{-4mm}  
						\\ 
						RNAN \hspace{-4mm} &
						RDN \hspace{-4mm} &
						SwinIR \hspace{-4mm} &
						Restormer \hspace{-4mm} &
						\textbf{Xformer (ours)} \hspace{-4mm}
						\\
					\end{tabular}
				\end{adjustbox}
				
		\end{tabular}}
		\vspace{-4mm}
		\caption{Visual comparisons on Gaussian color and gray image denoising ($\sigma$=50).}
		\label{fig:img_dn_visual}
		\vspace{-8mm}
	\end{figure*}
	
	\vspace{-3mm}
	\section{Experimental Results}
	\label{sec:experiment}
	\vspace{-2mm}
	\subsection{Experimental Settings}
	\noindent \textbf{Data and Evaluation.} We conduct experiments on two denoising tasks, including synthetic image denoising using noisy images generated with additive white Gaussian noise and real image denoising using real-world noisy images. For Gaussian denoising, following the previous work~\citep{swinir2021}, we use DIV2K~\citep{timofte2017ntire}, Flickr2K~\citep{lim2017enhanced}, BSD500~\citep{arbelaez2010contourbsd500}, and WED~\citep{ma2016waterloowed} as training data. Set12~\citep{zhang2017beyonddncnn}, BSD68~\citep{martin2001database}, Kodak24~\citep{franzen1999kodak}, McMaster~\citep{zhang2011color}, and Urban100~\citep{huang2015single} are the testing data. For real image denoising, same with Restormer~\citep{restormer2022}, we use SIDD~\citep{abdelhamed2018high} to train our model. The evaluation is performed on 1,280 patches of the SIDD validation set~\citep{abdelhamed2018high} and 50 pairs of images from the DND~\citep{plotz2017benchmarking}. Note that we evaluate the performance with the commonly-used PSNR and SSIM~\citep{wang2004image} metrics. Besides, we also provide comparisons of FLOPs and model size. We set the input image size to 3$\times$128$\times$128 when calculating FLOPs.
	
	\noindent \textbf{Training Settings.} We perform data augmentation on the training data through random horizontal or vertical flips and rotation of $90^{\circ}$, $180^{\circ}$, and $270^{\circ}$. Using progressive training strategy proposed by Restormer~\citep{restormer2022}, we set the batch size and patch size pairs to [(64,$128^2$), (40,$160^2$), (32,$192^2$), (16,$256^2$), (8,$320^2$), (8,$384^2$)] at training iterations [0k, 92k, 156k, 204k, 240k, 276k]. AdamW~\citep{loshchilov2017decoupled} is used to optimize our model with $\beta_1=0.9$, $\beta_2=0.999$, and weight decay $10^{-4}$. We train our model for total 300k iterations and the initial learning rate is set to 3$\times10^{-4}$ and gradually reduced to $10^{-6}$ through the cosine annealing~\citep{loshchilov2016sgdr}. Our Xformer is implemented on PyTorch~\citep{paszke2017automatic} using 4 Nvidia A100 GPUs.
	
	\vspace{-5mm}
	\subsection{Ablation Study} \label{sec:ablation}
	\vspace{-1mm}
	For ablation experiments, we train all the models on Gaussian color image denoising task with noise level $\sigma$=50. We train these models for 100k iterations. The evaluations are performed on Urban100 dataset. We also report the model size and FLOPs. The results are shown in Tab.~\ref{tab:ablations}.
	
	\noindent \textbf{Impact of STB and CTB.} We design the ablation study to support using both STB and CTB. 
	
	\noindent \emph{Ablation design.} As shown in Tab.~\ref{tab:STBCTB}, we present three different network designs. Specifically, using all STB or CTB means that we replace all the Transformer blocks in Xformer with STB or CTB. For fair comparisons, we design these models with comparable complexity.
	
	\noindent \emph{Analyses and conclusion.} As we can see, the model using all CTB gets poor performance since it pays less attention to patch-level information. Furthermore, the model using all STB obtains suboptimal performance. However, using STB and CTB together in dual branches can achieve the best performance gains. The joint application of hybrid Transformer blocks can simultaneously obtain patch-level and channel-level information, which is very important.
	
	\noindent \textbf{Impact of BCU.} We further discuss the impact of the BCU by the ablation study. 
	
	\noindent \emph{Ablation design.} We set four different experimental conditions. As shown in Tab.~\ref{tab:bcu}, we compare the results of models without BCU, using single-direction BCU, and using complete BCU. Note that using single-direction BCU means that we only use the DWConv or the Conv to provide the information transmission from a single direction. Furthermore, BCU-1 denotes the model using DWConv and BCU-2 denotes the model using Conv. 
	
	\noindent \emph{Analyze the importance of using BCU.} As we can see, the model with complete BCU achieves PSNR gain of 0.12 dB over that without BCU, which indicates that the BCU is an important component. 
	
	\noindent \emph{Analyze the necessity of the interactive manner.} The models using single-direction BCU have suboptimal performance. Moreover, we find that the model with BCU-2 yields 0.08 dB gain over that with BCU-1. It reveals that the information flow from the channel-wise branch has a bigger impact.
	
	\noindent \emph{Visual analyses and conclusion.} We provide visual results in Fig.~\ref{fig:feature visual}. We visualize the deep features from the last STB or CTB in corresponding branches. We can see that the STB and CTB in the network with BCU can capture more extra information and thus show better visual results. We conclude that our proposed BCU can provide effective information fusion for the dual branches. With BCU, our proposed network can achieve promising performance.
	
	\noindent \textbf{Impact of Different Branches.} We also discuss the importance of different branches.
	
	\noindent \emph{Ablation design.} We design the networks with comparable model size and FLOPs. As shown in Tab.~\ref{table:branch}, we present four different networks, including using STB-based branch, using CTB-based branch, using dual branches without BCU, and using dual branches with BCU. The model using single branch is degraded to a complete U-shaped network like the architecture of Restormer.
	
	\noindent \emph{Analyze the results of using single branch.} The model using STB-based branch performs better than that using CTB-based one. It demonstrates that the patch-level information deserves more attention. 
	
	\noindent \emph{Analyze the results of using dual branches without BCU.} We find that the model using dual branches without BCU gets limited performance. It confirms the statement we discuss in Sec.~\ref{subsec: bcu}. The simple concatenating operation fails to effectively utilize the obtained information. Thus, the direct connection of dual branches even brings unsatisfied results.
	
	\noindent \emph{Analyze the results of using dual branches with BCU.} As we can see, with BCU, our model using dual branches can achieve greatly enhanced performance. It further validates that the BCU plays an important role in fusing patch-level and channel-level information in our concurrent network.
	
	\noindent \emph{Conclusion.} Thanks to these two joint designs, our proposed method can explore enhancing global information modeling in Transformer and thus outperform previous promising methods.
	
	\noindent \textbf{Impact of Shift.} Table~\ref{tab:shift} shows the comparisons about whether to use shift. As we can see, the model using shift operation obtains better performance gains. It is because shift operation can bring more global receptive fields for window-based self-attention. Therefore, we choose to use shift.
	
	\begin{table*}[t]
		\centering
		\vspace{-1mm}
		\resizebox{1.0\textwidth}{!}{
			\setlength{\tabcolsep}{0.7mm}
			\centering
			\begin{tabular}{c|c|cccccccccccc} 
				\toprule[0.15em]
				\rowcolor{color3}	Dataset~&$~~\sigma$~~ &BM3D & DnCNN & IRCNN & FFDNet & NLRN & MWCNN & RNAN & RDN & DRUNet & SwinIR & Restormer & Xformer (ours) \\
				\midrule[0.15em]
				&15 & 32.37 & 32.86 & 32.76 & 32.75 & 33.16 & 33.15 & - & - & 33.25 & 33.36 & \underline{33.42} & \bf 33.46  \\
				Set12 &25 & 29.97 & 30.44 & 30.37 & 30.43 & 30.80 & 30.79 & - & - & 30.94 & 31.01 & \underline{31.08} & \bf 31.16  \\
				&50 & 26.72 & 27.18 & 27.12 & 27.32 & 27.64 & 27.74 & 27.70 & 27.60 & 27.90 & 27.91 & \underline{28.00} & \bf 28.10  \\
				\midrule
				&15 & 31.08 & 31.73 & 31.63 & 31.63 & 31.88 & 31.86 & - & - & 31.91 & \underline{31.97} & 31.96 & \bf 31.98 \\
				BSD68 &25 & 28.57 & 29.23 & 29.15 & 29.19 & 29.41 & 29.41 & - & - & 29.48 & 29.50 & \underline{29.52} & \bf 29.55  \\
				&50 & 25.60 & 26.23 & 26.19 & 26.29 & 26.47 & 26.53 & 26.48 & 26.41 & 26.59 & 26.58 & \underline{26.62} & \bf 26.65 \\
				\midrule
				& 15& 32.35 & 32.64 & 32.46 & 32.40 & 33.45 & 33.17 & - & - & 33.44 & 33.70 & \underline{33.79} & \bf 33.98 \\
				Urban100 & 25& 29.70 & 29.95 & 29.80 & 29.90 & 30.94 & 30.66 & - & - & 31.11 & 31.30 & \underline{31.46} & \bf 31.78 \\
				& 50& 25.95 & 26.26 & 26.22 & 26.50 & 27.49 & 27.42 & 27.65 & 27.40 & 27.96 & 27.98 & \underline{28.29} & \bf 28.71 \\
				\bottomrule[0.15em]
		\end{tabular}}
		\vspace{-2mm}
		\caption{\small PSNR (dB) comparisons for \underline{\textbf{Gaussian grayscale image denoising}} on three benchmark datasets. The \underline{underlined} and \textbf{bold} numbers indicate the second best and the best results.}\label{Tab:performance grayscale}
		\vspace{-2mm}
	\end{table*}
	
	\begin{table*}[t]
		\centering
		\resizebox{1.0\textwidth}{!}{
			\setlength{\tabcolsep}{0.7mm}
			\centering
			\begin{tabular}{c|c|ccccccccccccc} 
				\toprule[0.15em]
				\rowcolor{color3}	Dataset~&$~~\sigma$~~ &BM3D & DnCNN & IRCNN & FFDNet & RNAN & RDN & DRUNet & P3AN & IPT & SwinIR & Restormer & Xformer (ours)\\
				\midrule[0.15em]
				&15 & 33.52 & 33.90& 33.86& 33.87 & - & - & 34.30 & - & - & \underline{34.42} & 34.40 & \bf 34.43   \\
				CBSD68 &25 & 30.71 & 31.24 & 31.16 & 31.21 & - & - & 31.69 & -  & - & 31.78 & \underline{31.79} & \bf 31.82  \\
				&50 & 27.38 & 27.95 & 27.86 & 27.96 & 28.27 & 28.31 & 28.51 & 28.37 & 28.39 & 28.56 & \underline{28.60} & \bf 28.63 \\
				\midrule
				&15 & 34.28 & 34.60 & 34.69 & 34.63 & - & - & 35.31 & - & - & 35.34 & \textcolor{red}{*}\underline{35.35} & \bf 35.39 \\
				Kodak24 &25 & 32.15 & 32.14 & 32.18 & 32.13 & - & - & 32.89 & - & - & 32.89 & \textcolor{red}{*}\underline{32.93} & \bf 32.99 \\
				&50 & 28.46 & 28.95 & 28.93 & 28.98 & 29.58 & 29.66 & 29.86 & 29.69 & 29.64 & 29.79 & \textcolor{red}{*}\underline{29.87} & \bf 29.94 \\
				\midrule
				& 15& 34.06 & 33.45 & 34.58 & 34.66 & - & - & 35.40 & - & - & 35.61 & \underline{35.61} & \bf 35.68 \\
				McMaster & 25& 31.66 & 31.52 & 32.18 & 32.35 & - & - & 33.14 & - & - & 33.20 & \underline{33.34} & \bf 33.44 \\
				& 50& 28.51 & 28.62 & 28.91 & 29.18 & 29.72 & - & 30.08 & - & 29.98 & 30.22 & \underline{30.30} & \bf 30.38  \\
				\midrule
				& 15& 33.93 & 32.98 & 33.78 & 33.83 & - & - & 34.81 & - & - & 35.13 & \underline{35.13} & \bf 35.29 \\
				Urban100 & 25& 31.36 & 30.81 & 31.20 & 31.40 & - & - & 32.60 & - & - & 32.90 & \underline{32.96} & \bf 33.21 \\
				& 50& 27.93 & 27.59 & 27.70 & 28.05 & 29.08 & 29.38 & 29.61 & 29.51 & 29.71 & 29.82 & \underline{30.02} & \bf 30.36  \\
				\bottomrule[0.15em]
		\end{tabular}}
		\vspace{-1mm}
		\caption{\small PSNR (dB) comparisons for \underline{\textbf{Gaussian color image denoising}} on four benchmark datasets. The \underline{underlined} and \textbf{bold} numbers indicate the second best and the best results. \textcolor{red}{*} denotes results which are obtained by testing with officially provided pre-trained models.}\label{Tab:performance color}
		\vspace{-4mm}
	\end{table*}
	
	\begin{table*}[t]
		\centering
		\resizebox{1\textwidth}{!}{
			\setlength{\tabcolsep}{0.7mm}
			\centering
			\begin{tabular}{c|c|ccccccccccccccccc} 
				\toprule[0.15em] 
				\rowcolor{color3}
				Dataset & Method  &  BM3D & DnCNN & CBDNet  & RIDNet  & AINDNet & VDN & SADNet &DANet & CycleISP & MIRNet & DeamNet & DAGL & MAXIM &  Uformer & Restormer & Xformer \\
				\midrule[0.15em]
				SIDD &PSNR & 25.65 & 23.66& 30.78  &  38.71  &  39.08  & 39.28  & 39.46  & 39.47 & 39.52  & 39.72  & 39.47  & 38.94 & 39.96 & 39.89 & \textbf{40.02} & \underline{39.98}\\
				&SSIM & 0.685 & 0.583  & 0.801  &  0.951  &  0.954  & 0.956  & 0.957  & 0.957 & 0.957  & 0.959  & 0.957  & 0.953 & 0.960 & 0.960 & \underline{0.960} & \textbf{0.960}\\
				\midrule
				DND&PSNR & 34.51 & 32.43 & 38.06  &  39.26  &  39.37  & 39.38  & 39.59  & 39.58 & 39.56  & 39.88  & 39.63  & 39.77 & 39.84 & \underline{40.04} & 40.03 & \textbf{40.19}\\
				&SSIM & 0.851 & 0.790 & 0.942  &  0.953  &  0.951  & 0.952  & 0.952  & 0.955 & 0.956  & 0.956  & 0.953 & 0.956 & 0.954 & 0.956 & \underline{0.956} & \textbf{0.957} \\
				\bottomrule[0.15em]
		\end{tabular}}
		\vspace{-1mm}
		\caption{\small PSNR (dB) and SSIM comparisons for \underline{\textbf{real image denoising}} on two benchmark datasets. The \underline{underlined} and \textbf{bold} numbers indicate the second best and the best results.}\label{Tab:performance real}
		\vspace{-3mm}
	\end{table*}

	\vspace{-5mm}
	\subsection{Gaussian Image Denoising Results}
	\vspace{-2mm}
	We provide the comparisons of our Xformer with recent representative image denoising methods on both Gaussian color and grayscale image denoising tasks. As shown in Tabs.~\ref{Tab:performance grayscale} and Tab.~\ref{Tab:performance color}, BM3D~\citep{dabov2007imageBM3D} is the classical denoising method. The CNN-based methods include DnCNN~\citep{zhang2017beyonddncnn}, IRCNN~\citep{zhang2017learningIRCNN}, FFDNet~\citep{zhang2018ffdnet}, NLRN~\citep{liu2018nonNLRN}, MWCNN~\citep{liu2018multiMWCNN}, RNAN~\citep{zhang2019rnan}, RDN~\citep{zhang2020rdnir}, DRUNet~\citep{zhang2021plugDRUNet}, and P3AN~\citep{hu2021pseudoP3AN}. The Transformer-based methods include SwinIR~\citep{swinir2021}, IPT~\citep{chen2021preIPT}, and Restormer~\citep{restormer2022}. All the results are obtained by open available data. We get some results by using officially provided pre-trained models. Following most recent works~\citep{swinir2021,restormer2022}, we set the additional noise level to $15$, $25$, and $50$. We also provide the visual comparisons in Fig.~\ref{fig:img_dn_visual}. The results of model parameters and FLOPs comparisons are shown in Tab.~\ref{table:complexity}.
	
	\vspace{0.4em}
	\noindent \textbf{Quantitative Comparisons.} We present the PSNR results of all compared approaches in Tab.~\ref{Tab:performance grayscale} and Tab.~\ref{Tab:performance color}. The corresponding scores are obtained by testing on several benchmark datasets for Gaussian grayscale and color image denoising. As we can see, our Xformer achieves the best PSNR performance across all evaluation datasets. Specifically, for the evaluation on high-resolution Urban100 dataset~\citep{huang2015single} under the challenging noise level 50, Xformer obtains 0.42 dB performance gain over the previous best Transformer-based network Restormer~\citep{restormer2022}, as shown in Tab.~\ref{Tab:performance grayscale}. Similarly, Table~\ref{Tab:performance color} also shows that the best performance is achieved by Xformer for Gaussian color image denoising. Compared to the results of SwinIR~\citep{swinir2021}, Xformer has better performance gains while maintaining 4.76$\times$ fewer FLOPs. It is also worth mentioning that our proposed method has the comparable model size and FLOPs with Restormer. In short, the experimental results demonstrate that our proposed Xformer becomes a new promising Transformer-based network for the Gaussian image denoising.
	
	\vspace{0.4em}
	\noindent \textbf{Visual Comparisons.} The visual comparisons for Gaussian color and grayscale image denoising on some challenging examples are shown in Figs.~\ref{fig:img_dn_visual}. The noise level is set to 50 and results are obtained by testing on Urban100. We can see that our Xformer is able to remove heavy noise corruption for color image denoising. Compared to some previous denoising methods, our method obtains visually pleasing results. Besides, for the gray image denoising, the detailed textures and high-frequency components of the original images are reserved by using our method. However, others suffer from the heavy blurring and missing details. It demonstrates that our Xformer performs excellently for both color and grayscale image denoising for visual results.
	
	\vspace{0.2em}
	\noindent \textbf{Model Size Comparisons.} Table~\ref{table:complexity} provides comparisons of parameters number and FLOPs with existing state-of-the-art methods. We calculate the FLOPs assuming that the input size is 3$\times$128$\times$128. The PSNR scores are reported on the popular benchmark datasets McMaster~\citep{zhang2011color} and Urban100~\citep{huang2015single} under Gaussian color image denoising with noise level $\sigma$=50. We mainly compare our proposed method to recent Transformer-based networks, including SwinIR and Restormer. We find that our Xformer enjoys very low FLOPs when compared to SwinIR. Meanwhile, it is seen that our method has comparable model size and FLOPs with Restormer. However, our Xformer can achieve the best performance among them. In special, it obtains 0.34 dB higher PSNR score over Restormer on Urban100. It indicates that our method has an acceptable computational and memory cost while performing promising image denoising.

	\vspace{-1mm}
	\subsection{Real Image Denoising Results}
	As shown in Tab.~\ref{Tab:performance real}, We show the quantitative comparisons of our Xformer with other state-of-the-art methods on real-world image denoising task. In detail, we report the evaluation results from the classic denoising method BM3D~\citep{dabov2007imageBM3D}, CNN-based methods DnCNN~\citep{zhang2017beyonddncnn}, CBDNet~\citep{guo2019towardCBDNet}, RIDNet~\citep{anwar2019realRIDNet}, AINDNet~\citep{kim2020transferAINDNet}, VDN~\citep{yue2019variationalVDN}, SADNet~\citep{chang2020spatialSADNet}, DANet~\citep{yue2020dual}, CycleISP~\citep{zamir2020cycleisp}, MIRNet~\citep{zamir2020learningMIRNet}, DeamNet~\citep{ren2021adaptiveDeamNet}, DAGL~\citep{mou2021dynamicDAGL}, MAXIM~\citep{tu2022maxim}, and Transformer-based methods Uformer~\citep{wang2022uformer} and Restormer~\citep{restormer2022}. Following the recent work~\citep{restormer2022}, we only use SIDD~\citep{abdelhamed2018high} dataset to train our models. Then, the trained models are directly used to perform evaluations on the DND~\citep{plotz2017benchmarking} benchmark. As DND does not provide ground-truth labels, the corresponding results are obtained by uploading images to the online server. Note that all the results are obtained from the open-source data.
	
	\vspace{0.4em}
	\noindent \textbf{Quantitative Comparisons}. Table~\ref{Tab:performance real} shows the PSNR and SSIM scores of recent approaches on real-world image denoising. As we can see, our proposed Xformer outperforms all the state-of-the-art methods on the DND dataset~\citep{plotz2017benchmarking} and achieves comparable performance on the SIDD dataset~\citep{abdelhamed2018high}. Compared to all the CNN-based methods, our Xformer has the obvious performance improvement. Furthermore, compared to Restormer~\citep{restormer2022}, our method performs better with comparable mode complexity. Besides, our Xformer achieves higher performance gains than Uformer~\citep{wang2022uformer} while maintaining 2.01$\times$ fewer model parameters. In special, Restormer only utilized channel-wise self-attention and paid less attention to local patches interactions. On the contrary, Uformer focused on spatial-wise self-attention and neglected the channel-dimension token interactions. In contrast, our proposed Xformer explore building interactions among tokens from both spatial and channel dimensions. Through fusing both patch-level and channel-level information, our method is able to obtain better performance.
	
	\section{Conclusion}
	In this work, we propose a hybrid X-shaped vision Transformer, named Xformer, for image denoising task. We design a concurrent network structure to utilize spatial-wise window-based Transformer blocks and channel-wise Transformer blocks respectively in two branches. Our proposed Xformer can enable each branch to proceed representation learning from the corresponding dimension, spatial or channel. Besides, we propose the Bidirectional Connection Unit (BCU) to bridge the separate branches. Specifically, the BCU provides information fusion in an interactive manner and greatly enhances the global information modeling ability of both branches. We conduct extensive experiments on the synthetic and real-world image denoising tasks. Experimental results demonstrate that our Xformer can outperform recent state-of-the-art methods both quantitatively and visually with comparable model size and computational cost.
	
	\section*{Acknowledgments} 
	This work was supported in part by NSFC grant (62141220, U19B2035) and Shanghai Municipal Science and Technology Major Project (2021SHZDZX0102).
	
	\bibliography{iclr2024_conference}
	\bibliographystyle{iclr2024_conference}
	
\end{document}